\documentclass[runningheads]{llncs}
\usepackage{graphicx}
\usepackage{amsmath,amssymb} 
\usepackage{color}
\usepackage{tabularx}
\usepackage{multirow}
\usepackage{pifont}
\usepackage[FIGTOPCAP,scriptsize]{subfigure}

\usepackage[super]{nth}
\usepackage{booktabs}
\usepackage{wrapfig}

\newcommand{\etal}{\textit{et al.}}
\newcommand{\eg}{\textit{e.g.}}
\newcommand{\ie}{\textit{i.e.}}

\begin{document}
\pagestyle{headings}
\mainmatter

\def\ACCV20SubNumber{505}  

\title{Learning 3D Face Reconstruction with \\ a Pose Guidance Network} 
\titlerunning{Learning 3D Face Reconstruction with a Pose Guidance Network}
%
\author{Pengpeng Liu\inst{1}\thanks{Work mainly done during an internship at Huya AI.} \and
Xintong Han\inst{2} \and
Michael Lyu\inst{1} \and
Irwin King\inst{1} \and
Jia Xu\inst{2}
}
\authorrunning{P. Liu et al.}
%
\institute{The Chinese University of Hong Kong \and
Huya AI\\}

\maketitle

\begin{abstract}
We present a self-supervised learning approach to learning monocular 3D face reconstruction with a pose guidance network (PGN). First, we unveil the bottleneck of pose estimation in prior parametric 3D face learning methods, and propose to utilize 3D face landmarks for estimating pose parameters. With our specially designed PGN, our model can learn from both faces with fully labeled 3D landmarks and unlimited unlabeled in-the-wild face images. Our network is further augmented  with a self-supervised learning scheme, which exploits face geometry information embedded in multiple frames of the same person, to alleviate the ill-posed nature of regressing 3D face geometry from a single image. These three insights yield a single approach that combines  the complementary strengths of parametric model learning and data-driven learning techniques. We conduct a rigorous evaluation on the challenging AFLW2000-3D, Florence and FaceWarehouse datasets, and show that our method outperforms the state-of-the-art for all metrics.
\end{abstract}

\section{Introduction}

\label{sec:intro}
Monocular 3D face reconstruction with precise geometric details serves as a foundation to a myriad of computer vision and graphics applications, including  face recognition \cite{blanz2003face,paysan20093d}, digital avatars \cite{nagano2018pagan,hu2017avatar}, face manipulation \cite{thies2016face2face,kim2018deep}, {\it etc}. However, this problem is extremely challenging due to its ill-posed nature, as well as difficulties to acquire accurate 3D face annotations.

Most successful attempts to tackle this problem are built on parametric face models, which usually contain three sets of parameters: identity, expression, and pose. The most famous one is 3D Morphable Model (3DMM)~\cite{blanz1999morphable} and its variants \cite{thies2016face2face,saito2016real,cao2014displaced,cao20133d}. Recently, CNN-based methods that directly learn to regress the parameters of 3D face models \cite{sanyal2019learning,yi2019mmface,chang2018expnet,wu2019mvf},  achieve state-of-the-art performance.

\begin{figure}[t]
\centering
\includegraphics[width=0.9\textwidth]{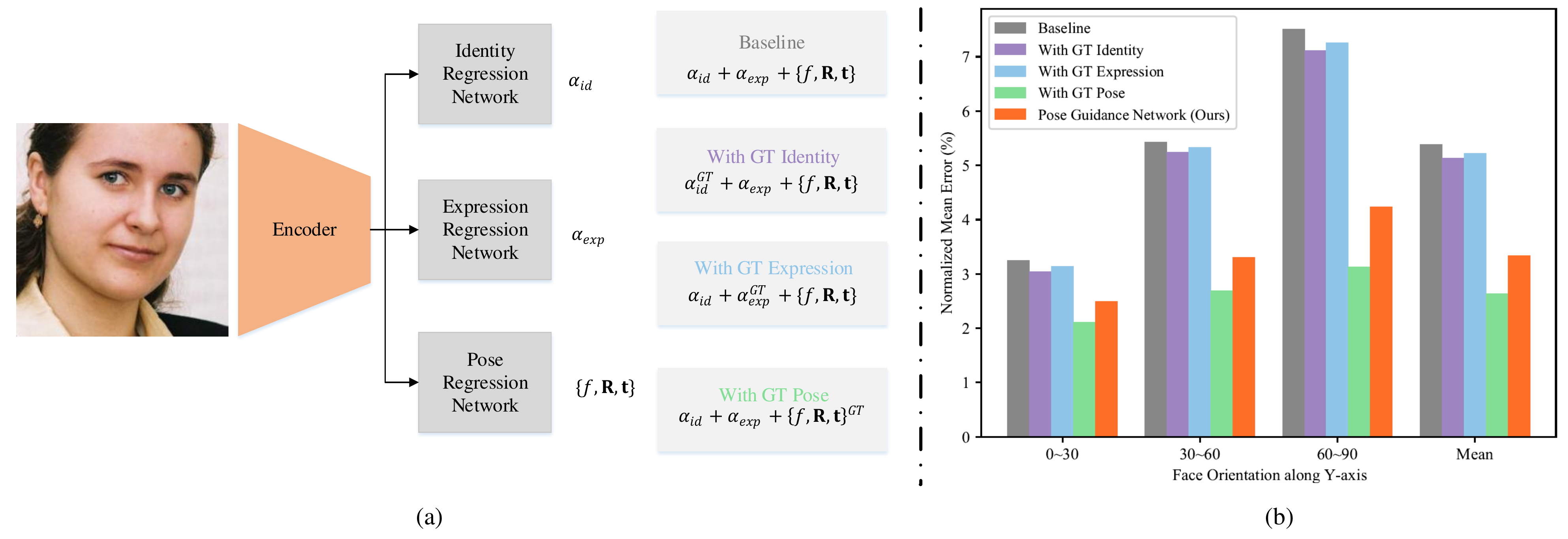}
\caption{Our CNN baseline takes an RGB image as input, and regresses identity, expression and pose parameters simultaneously. The three sets of parameters are obtained by minimizing the 3D vertex error. We compute the Normalized Mean Error (NME) of this face model and denote it as \texttt{Baseline}.
Then we replace  the predicted identity, expression, pose parameters with their ground truth, and recompute the NME respectively: \texttt{With GT Identity, Expression, Pose}. As shown in (b), \texttt{With GT Pose} yields the highest performance gain, and the gain is more significant as the face orientation degree increases. Our \texttt{Pose Guidance Network} takes advantage of this finding (Sec. \ref{sec:pgn}), and greatly reduces the error caused by inaccurate pose parameter regression.}
\label{teaser}
\end{figure}

Are these  parameters well disentangled and can they be accurately regressed by CNNs?
To answer this question, we conduct a careful study  on the AFLW2000-3D dataset \cite{zhu2016face}.
Fig.~\ref{teaser}(a) illustrates our setting. We first train a neural network that takes an RGB image as input to simultaneously regress the identity, expression and pose parameters.  The \texttt{Baseline}  3DMM model is obtained by minimizing the 3D vertex error.
Then, we independently replace the predicted identity, expression, and pose parameters with their corresponding ground truth parameters (denoted as \texttt{GT Identity}, \texttt{GT Expression}, and \texttt{GT Pose}), and  recompute the 3D face reconstruction error shown in Fig.~\ref{teaser}(b).

Surprisingly, we found that \texttt{GT Pose} yields almost 5 times more performance gain than its two counterparts.
The improvement is even more significant when the face orientation degree increases. We posit that there are two reasons causing this result: (1)
These three sets of parameters are heavily correlated, and predicting a bad pose will dominate the identity and expression estimation of the 3D face model;
(2) 3D face annotations are scarce especially for those with unusual poses.

To address these  issues, we propose  a pose guidance network (PNG) to
isolate the pose estimation from the original 3DMM parameters regression by estimating a UV position map \cite{feng2018joint} for 3D face landmark vertices.
Utilizing the predicted 3D landmarks help to produce more accurate face poses compared to joint parameters regression (\ie, \texttt{Baseline} in Fig.~\ref{teaser}), and the predicted 3D landmarks also contain valuable identity and expression information that further refines the estimation of identity and expression. Moreover, this enables us to learn from both accurate but limited 3D annotations, and unlimited in-the-wild images with pseudo 2D landmarks (from off-the-shelf landmark extractor like \cite{bulat2017far}) to predict more accurate 3D landmarks. Consequently, with our proposed PGN, the performance degradation brought by inaccurate pose parameter regression is significantly mitigated as shown in Fig.~\ref{teaser}(b).

To further overcome the scarcity of 3D face annotations, we leverage the readily available in-the-wild videos by introducing a novel set of self-consistency loss functions to boost the performance. Given 3D face shapes in multiple frames of the same subject, we render a new image for each frame by replacing its texture with that of \textit{commonly visible vertices} from other images. Then, by forcing the rendered image to be consistent with the original image in photometric space, optical flow space and semantic space, our network learns to avoid depth ambiguity and predicts better 3D shapes even without explicitly modeling albedo.

We summarize our key contributions as follows:

(1) We propose a PGN to solely predict the 3D landmarks for estimating the pose parameters based on a careful study (Fig. \ref{teaser}). The PGN effectively reduces the error compared to directly regressing the pose parameters and provides informative priors for 3D face reconstruction.

(2) The PGN allows us to utilize both fully annotated 3D landmarks and pseudo 2D landmarks from unlabeled in-the-wild data. This leads to a more accurate landmark estimator and thus helping better 3D face reconstruction.

(3) Built on a visible texture swapping module, our method explores multi-frame shape and texture consistency in a self-supervised manner, while carefully handling the occlusion and illumination change across frames.

(4) Our method shows superior qualitative and quantitative results on ALFW-2000-3D~\cite{zhu2016face}, Florence~\cite{bagdanov2011florence} and FaceWarehouse~\cite{cao2013facewarehouse} datasets.

\section{Related Work}
Most recent 3D face shape models are derived from Blanz and Vetter 3D morphable models (3DMM) \cite{blanz1999morphable}, which represents 3D faces with linear combination of PCA-faces from a collection of 3D face scans. To make 3DMM more representative, Basel Face Model (BFM) \cite{paysan20093d} improved  shape and texture accuracy, and FaceWarehouse~\cite{cao2013facewarehouse} constructed a set of individual-specific expression blend-shapes. Our approach  is also built on 3DMM --- we aim to predict 3DMM parameters to reconstruct 3D faces from monocular frames.

\textbf{3D Face Landmark Detection and Reconstruction.} 3D face landmark detection and 3D face reconstruction are closely related. On the one hand, if the 3DMM parameters can be estimated accurately, face landmark detection can be greatly improved, especially for the occluded landmarks~\cite{zhu2016face}. Therefore, several approaches~\cite{zhu2016face,liu2017dense,gou2016shape} aligned 3D face by fitting a 3DMM model. On the other hand, if 3D face landmarks are precisely estimated, it can provide strong guidance for 3D face reconstruction. Our  method goes towards the second direction---we first estimates 3D face landmarks by regressing UV position map and then utilizes it to guide 3D face reconstruction.

\textbf{3D Face Reconstruction from a Single Image.} To reconstruct 3D faces from a single image, prior methods~\cite{thies2016face2face,blanz2003face,romdhani2005estimating} usually conduct iterative optimization methods to fit 3DMM models by leveraging facial landmarks or local features \eg, color or edges. However, the convergence of optimization is very sensitive to the initial parameters.
Tremendous progress has been made by CNNs that directly regress 3DMM parameters~\cite{zhu2016face,dou2017end,tuan2017regressing}.
Jackson~\etal~\cite{jackson2017large} directly regressed the full 3D facial structure via volumetric convolution.
Feng~\etal~\cite{feng2018joint} predicted a UV position map to represent the full 3D shape. MMFace~\cite{yi2019mmface} jointly trained a volumetric network and a parameter regression network, where the former one is employed to refine pose parameters with ICP as a post-processing.
All these three methods need to be trained in a supervised manner, requiring full 3D face annotations, which are limited at scale~\cite{zhu2016face}. To bypass the limitation of training data, Tewari~\etal~\cite{tewari2017mofa} and Genova~\etal~\cite{genova2018unsupervised} proposed to fit 3DMM models with only unlabeled images. They show that it is possible to achieve great face reconstruction in an  unsupervised manner by minimizing photometric consistency or facial identity loss. Later, Chang~\etal~\cite{chang2019deep} proposed to regress identity, expression and pose parameters with three independent networks. However, due to depth ambiguity, these unsupervised monocular methods fail to capture precise 3D facial structure.
In this paper, we propose to mitigate the limitation of datasets by utilizing both labeled and unlabeled datasets, and to learn better facial geometry from multiple frames.

\textbf{3D Face Reconstruction from Multiple Images.}
Multiple images of the same person contain rich information for learning better 3D face reconstruction.
Piotraschke~\etal~\cite{piotraschke2016automated} introduced an automated algorithm that selects and combines reconstructions of different facial regions from multiple images into a single 3D face. RingNet~\cite{sanyal2019learning} considered shape consistency across different images of the same person, while we focus on face reconstruction from videos, where photometric consistency can be well employed. MVF~\cite{wu2019mvf} regressed 3DMM parameters from multi-view images. However, MVF assumes that the expressions in different views are the same, therefore its application is restricted to multi-view images. Our method does not have such constraint and can be applied to both single-view and multi-view 3D face reconstruction. 

The approach that is closest to ours is FML~\cite{tewari2019fml}, which learns face reconstruction from monocular videos by ensuring consistent shape and appearance across frames. However, it only adds multi-frame identity consistency constraints, which does not fully utilize geometric constraints among different images. Unlike FML, we do not model albedo to estimate texture parameters, but directly sample textures from images, swap commonly visible texture and project them onto different image planes while enforcing photometric and semantic consistency. Additionally, we introduce a PGN, which removes the need of pose parameter estimation and enables our model to produce more accurate identity and expression estimation.

\begin{figure*}[t]
\centering
\includegraphics[width=0.95\textwidth]{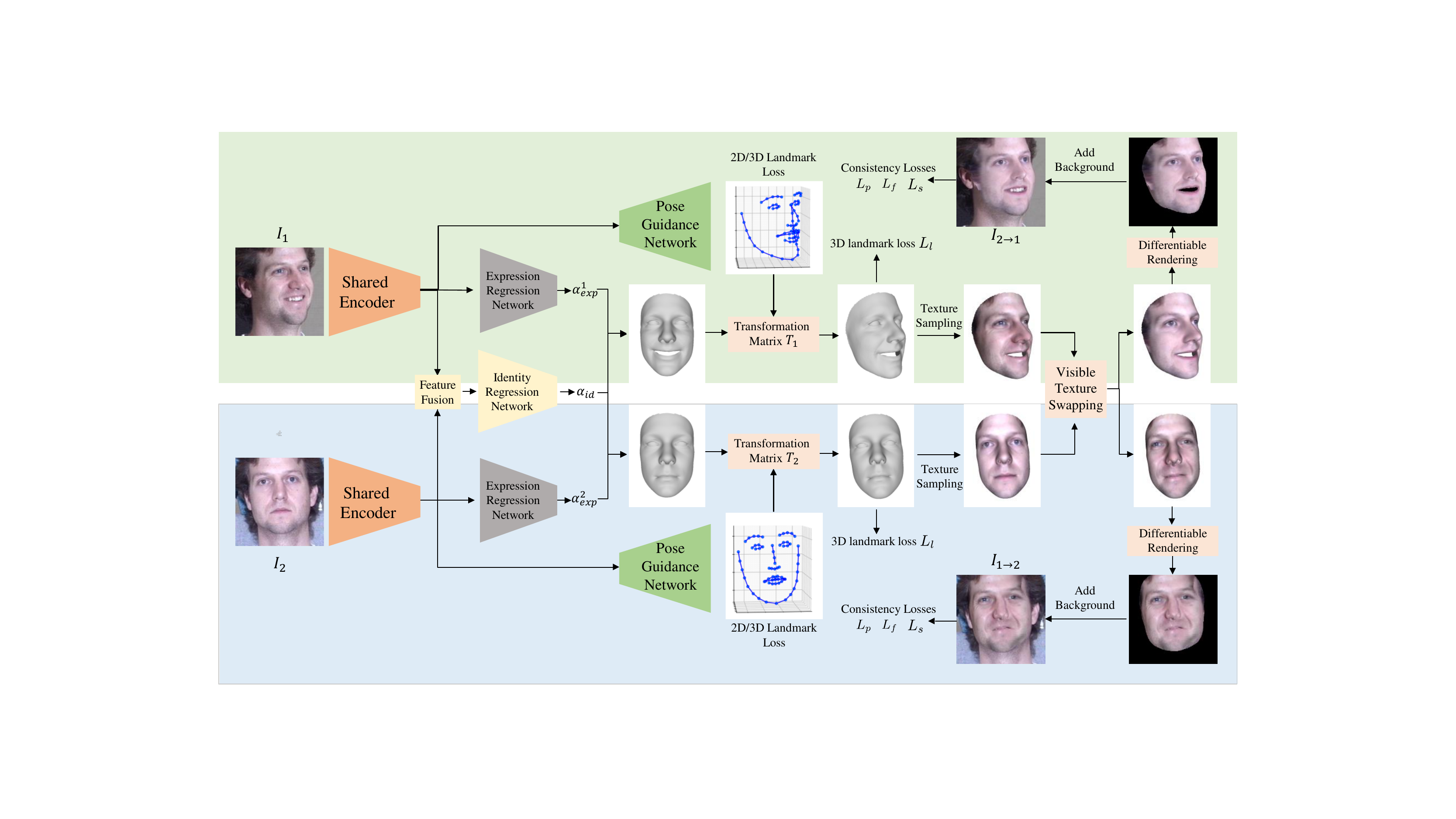}
\caption{\textbf{Framework overview.} Our shared encoder extracts semantic feature representation from multiple images of the same person. Then, our identity and expression regression networks regress 3DMM face identity and expression parameters (Sec.~\ref{sec:3DMM}) with accurate guidance of our PGN that predicts 3D face landmarks (Sec.~\ref{sec:pgn}). Finally, We utilize multiple frames  (Sec.~\ref{sec:multi-image}) to train our  proposed network with a set of self-consistency loss  functions (Sec.~\ref{sec:loss}).}
\label{Framework}
\end{figure*}

\section{Method}
We illustrate our framework overview in Fig.~\ref{Framework}. First, we utilize a shared encoder to extract semantic feature representations from multiple frames of the same person. Then, an identity regression branch and an expression regression branch are employed to regress 3DMM face identity and expression parameters (Sec. \ref{sec:3DMM}) with the help of our PGN that predicts 3D face landmarks (Sec. \ref{sec:pgn}). Finally, we explore self-consistency (Sec. \ref{sec:multi-image}) with our newly designed consistency losses (Sec. \ref{sec:loss}).

\subsection{Preliminaries}
\label{sec:3DMM}

Let $\textbf{S} \in \mathbb{R}^{3N}$ be a 3D face with $N$ vertices, $\overline{\textbf{S}} \in \mathbb{R}^{3N} $ be the mean face geometry, $\textbf{B}_{id} \in \mathbb{R}^{3N \times 199} $ and $\textbf{B}_{exp} \in \mathbb{R}^{3N \times 29}$ be PCA basis of identity and expression, $\boldsymbol{\alpha}_{id} \in \mathbb{R}^{199}$ and $\boldsymbol{\alpha}_{exp} \in \mathbb{R}^{29}$ be the identity and expression parameters. The classical 3DMM face model~\cite{blanz1999morphable} can be defined as follows:
\begin{equation}
\textbf{S}(\boldsymbol{\alpha}_{id}, \boldsymbol{\alpha}_{exp}) = \overline{\textbf{S}} + \textbf{B}_{id}\boldsymbol{\alpha}_{id} + \textbf{B}_{exp}\boldsymbol{\alpha}_{exp}.
\label{eq:3dmm}
\end{equation}
Here, we adopt BFM~\cite{paysan20093d} to obtain $\overline{\textbf{S}}$ and $\textbf{B}_{id}$, and expression basis $\textbf{B}_{exp}$ is extracted from FaceWareHouse~\cite{cao2013facewarehouse}. Then, we employ a  perspective projection model to project a 3D face point $\textbf{s}$ onto an image plane:
\begin{equation}
\textbf{v}(\boldsymbol{\alpha}_{id}, \boldsymbol{\alpha}_{exp}) = \left[\begin{matrix} 1 & 0 & 0 \\ 0 & 1 & 0 \end{matrix} \right] \cdot (f \cdot \textbf{R} \cdot \textbf{s} + \textbf{t}) = \left[\begin{matrix} 1 & 0 & 0 \\ 0 & 1 & 0 \end{matrix} \right] \cdot \left[\begin{matrix} f \cdot \textbf{R} & \textbf{t} \end{matrix} \right] \cdot \left[\begin{matrix} \textbf{s} \\ 1 \end{matrix} \right],
\label{eq:projetion}
\end{equation}
where $\textbf{v}$ is the projected point on the image plane, $f$ is a scaling factor, $\textbf{R}\in \mathbb{R}^{3\times3}$ indicates a rotation matrix, $\textbf{t} \in \mathbb{R}^{3}$ is a translation vector.

However, it is challenging for neural networks to regress identity parameter $\boldsymbol{\alpha}_{id}$, expression parameter $\boldsymbol{\alpha}_{exp}$ and pose parameter $\{f, \textbf{R}, \textbf{t}\}$ together, because these parameters cannot be easily disentangled and pose parameters turn to dominate the optimization, making it more difficult to estimate accurate identity and expression (as discussed in Sec. \ref{sec:intro} and illustrated in Fig. \ref{teaser}).

To address this issue, we design a robust landmark-based PGN to obtain the transformation matrix $\textbf{T} = \left[\begin{matrix} f \cdot \textbf{R} & \textbf{t} \end{matrix} \right] $ instead of directly regressing its parameters. Next, we describe our PGN in detail.

\subsection{Pose Guidance Network}
\label{sec:pgn}

To decouple the optimization of pose parameter $\{f, \textbf{R}, \textbf{t}\}$ with
identity parameter $\boldsymbol{\alpha}_{id}$ and expression parameter $\boldsymbol{\alpha}_{exp}$,
we design a multi-task architecture with two output branches (shown in Fig. \ref{Framework}).
One branch optimizes the traditional 3DMM identity   and expression parameters $\boldsymbol{\alpha}_{id}, \boldsymbol{\alpha}_{exp}$.
The other branch is trained to estimate a UV position map \cite{feng2018joint} for 3D face landmarks, which provide key guidance for pose estimation.

Specifically, Let $\textbf{X}$ be the 3D landmark positions in the face geometry $\textbf{S}$, and $\textbf{X}_{UV} $ be the 3D landmarks estimated from our UV position map decoder, we estimate a transformation matrix $\textbf{T}$ by,
\begin{equation}
\min\limits_{\textbf{T}}||\textbf{T} \cdot \left[\begin{matrix} \textbf{X}    \\ \textbf{1} \end{matrix} \right] - \textbf{X}_{UV} ||_2.
\end{equation}
Here, $\textbf{T}$ has a closed-form solution:
\begin{equation}
\textbf{T} = \textbf{X}_{UV}  \cdot \left[\begin{matrix} \textbf{X}    \\ \textbf{1} \end{matrix} \right]^T
        \cdot \Big(\left[\begin{matrix} \textbf{X}    \\ \textbf{1} \end{matrix} \right] \cdot \left[\begin{matrix} \textbf{X}    \\ \textbf{1} \end{matrix} \right]^T\Big)^{-1}.
\end{equation}
As a result, we convert the estimation of $\textbf{T}$ into the estimation of a UV position map for 3D face landmarks rather than regressing $\textbf{T}$'s parameters. This disentangles the pose estimation and results in better performance than joint regression of $\boldsymbol{\alpha}_{id}, \boldsymbol{\alpha}_{exp}$and $\{f, \textbf{R}, \textbf{t}\}$. Another merit of this design is enabling us to train our network with two types of images: images with  3D landmark annotations and in-the-wild unlabeled images with 2D facial landmarks  extracted by off-the-shelf detectors. During training, we sample one image batch with 3D landmark labels and another image batch from unlabeled datasets. 3D landmark loss and 2D landmark loss are minimized for them, respectively. For 3D landmarks, we calculate the loss across all $x$, $y$ and $z$ channels of the UV position map, while for 2D landmark loss, only $x$ and $y$ channels are considered. More abundant training data leads to more accurate pose estimation, and hence better face reconstruction.

Note our work is different from PRN~\cite{feng2018joint}, which utilizes a CNN to regress dense UV position maps for all 3D face points.
PRN requires dense 3D face labels which are extremely difficult to obtain. Our network learns directly from sparse landmark annotations, which are much easier to obtain and more accurate than the synthetic data derived from facial landmarks.

\subsection{Learning from Multiple Frames}
\label{sec:multi-image}
The PGN combined with identity and expression parameters regression can achieve quite accurate 3D face reconstruction, but the estimated 3D mesh lacks facial details.
This is because 3D landmarks can only provide a coarse prediction of identity and expression. To generate meshes with finer details, we leverage multi-frame images from monocular videos as input and explore their inherent complementary information. In contrast to the common perspective that first estimates albedo maps and then enforces photometric consistency~\cite{tewari2019fml}, we propose a self-consistency framework based on a visible texture swapping scheme.

Every vertex in a 3DMM model has a specific semantic meaning. Given multiple images with the same identity, we can generate one 3D mesh for every image. Every corresponding vertex of different meshes share the same semantic meaning, even though these images are captured with different poses, expressions, lightings, \textit{etc}. If we sample texture from one image and project it onto the second image that has different pose and expression, the rendered image should have the same identity, expression and pose as the second image despite the illumination change. Our multi-image 3D reconstruction is built on this intuition. 

More specifically, our method takes multiple frames of the same subject as input, and estimates the same set of identity parameters for all images, and different expressions and poses (obtained from 3D face landmarks output by our PGN) for each image. To generate the same identity parameters, we adopt a similar strategy as~\cite{tewari2019fml}, which fuses feature representations extracted from the shared encoders of different images via average pooling (Feature Fusion in Fig. \ref{Framework}). In this way, we can achieve both single-image and multi-image face reconstruction.

For simplicity, we assume there are two images of the same person as input (the framework can easily extend to more than two images), denoted as $I_1$ and $I_2$ respectively. Then, as illustrated on the left side of Fig. \ref{Framework}, we can generate two 3D meshes with the same identity parameter $\boldsymbol{\alpha}_{id}$, two different expression parameters $\boldsymbol{\alpha}_{exp}^1, \boldsymbol{\alpha}_{exp}^2$, and pose transformation matrices $\textbf{T}_1, \textbf{T}_2$ obtained by our PGN. After that, we sample two texture maps $C_1, C_2$ with Equation~\ref{eq:projetion}, and project the first texture $C_1$ onto the second image $I_2$ with its expression parameter $\boldsymbol{\alpha}_{exp}^2$ and pose transformation matrix $\textbf{T}_2$ to obtain rendered image $I_{1 \to 2}$. Similarly, we can project $C_2$ to $I_1$ to obtain the rendered image $I_{2 \to 1}$. Ideally, if there is no illumination change, $I_2$ shall be the same as $I_{1 \to 2}$ over their non-occluded facial regions. However, there exists occlusion and illumination usually changes a lot for different images in real-world scenarios. To this end, we introduce several strategies to overcome these issues.

\textbf{Occlusion Handling.} We adopt a simple strategy to effectively determine if a pixel is occluded or non-occluded based on triangle face normal direction. Given a triangle with three vertices, we can compute its normal $\textbf{n}=(n_x, n_y, n_z)$. If the normal direction towards outside of the face mesh (\ie, $n_z > 0$), we regard these three vertices as non-occluded; otherwise they are occluded. According to this principle, we can compute two visibility maps $M_1$ and $M_2$, where value 1 indicates the vertex is non-occluded and 0 otherwise. A common visibility map $M_{12}$ is then defined as:
\begin{equation}
M_{12}=M_1 \odot M_2,
\end{equation}
where value 1 means that the vertex is non-occluded for both 3D meshes.

Considering the occlusion, when projecting $C_1$ onto the second image, we combine $C_1$ and $C_2$ by
\begin{equation}
C_{1 \to 2} = C_1 \odot M_{12} + C_2 \odot (1 - M_{12}).
\end{equation}
That is, we alleviate the influence of the occlusion by only projecting the \textit{commonly visible texture} from $I_1$ to $I_2$ to generate $C_{1 \to 2}$, while \textit{keeping the original pixels for the occluded part}. In this way, the rendered image $I_{1 \to 2}$ shall have the same identity, pose and expression information as $I_2$. The projection from $I_2$ to $I_1$ can be derived in the same manner. 

\textbf{Illumination Change.} The sampled texture is not disentangled to albedo, lighting, etc. Due to lighting and exposure changes, even if we can estimate accurate 3D geometry, the rendered texture $I_{1 \to 2}$ is usually different from $I_2$. To cope with these issues, we propose three schemes. First, we adopt the Census Transform \cite{hafner2013census} from optical flow estimation, which has been shown to be very robust to illumination change when computing photometric difference (Eq. \ref{eqn:photo}). Specifically, we apply a 7$\times$7 census transform and then compute the Hamming distance between the reference image $I_2$ and the rendered image $I_{1 \to 2}$. Second, we employ an optical flow estimator~\cite{liu2019ddflow} to compute the flow between $I_2$ and the rendered image $I_{1 \to 2}$. Since optical flow provides a 2D dense correspondence constraint, if the face is perfectly aligned, the optical flow between $I_2$ and $I_{1 \to 2}$ should be zeros for all pixels, so we try to minimize difference, \ie, minimize the magnitude of optical flow between them (Eq. \ref{eqn:flow}). Third, even though illumination changes, the identity, expression and pose shall be the same for $I_2$ and $I_{1 \to 2}$. Therefore, they must share similar semantic feature representation. Since our shared encoder can extract useful information to predict facial landmarks, identity and expression parameters, we use it as a semantic feature extractor and compare the feature difference between $I_2$ and $I_{1 \to 2}$ (Eq. \ref{eqn:semantic}).

\subsection{Training Loss}
\label{sec:loss}

To train our network for accurate 3D face reconstruction, we define a set of self-consistency loss functions, and minimize the following combination:
\begin{equation}
L = L_{l} + L_{p} + L_{f} + L_{s} + L_{r}.
\label{equ:loss}
\end{equation}
Each loss term is defined in detail as follows. Note that for simplicity, we only describe these loss terms regarding projecting $I_1$ to $I_2$ (\ie, $I_{1 \to 2}$) and the other way around ($I_{2 \to 1}$) can be defined similarly.

\textbf{Sparse Landmark Loss.} Our landmark loss measures the difference between the landmarks of transformed face geometry $\textbf{T} \cdot \textbf{X}   $ and the prediction of PGN $\textbf{X}_{UV} $:
\begin{equation}
L_{l} = \lambda_{l} \sum |\textbf{T} \cdot \textbf{X}    - \textbf{X}_{UV} |
\end{equation}
This is the core guidance loss, which is trained with both 3D and 2D landmarks.

\textbf{Photometric Consistency Loss}. Photometric loss measures the difference between the target image and the rendered image over those visible regions. We can obtain the visible mask $M^{2d}$ on the image plane with differentiable mesh render \cite{genova2018unsupervised}. Note that $M^{2d}$ is different from the vertex visibility map $M$, where the former denotes whether the pixel is occluded on the image plane, and the latter denotes whether the vertex in 3D mesh is occluded.
Besides, considering that most of the face regions have very similar color, we apply a weighted mask $W$ to the loss function, where we emphasize eye, nose, and mouth regions with a larger weight of 5, while the weight is 1 for other face regions~\cite{feng2018joint}. The photometric loss then writes:
\begin{equation}
L_{p} = \lambda_{p} \frac{\sum \text{Hamming}|\text{Census}(I_2) - \text{Census}(I_{1 \to 2})| \odot M^{2d}_2 \odot W} {\sum M^{2d}_2 \odot W},
\label{eqn:photo}
\end{equation}
where $\texttt{Census}$ represents the census transform, $\texttt{Hamming}$ denotes Hamming distance, and $M^{2d}_{2}$ is the corresponding visibility mask.

\textbf{Flow Consistency Loss.} We use optical flow to describe the dense correspondence between the target image and the rendered image, then the magnitude of optical flow is minimized to ensure the visual consistency between two images:
\begin{equation}
L_{f} = \lambda_{f} \sum {|\textbf{w}(I_2, I_{1 \to 2})| \odot W} / \sum{W},
\label{eqn:flow}
\end{equation}
where $\textbf{w}$ is the optical flow computed from \cite{liu2019ddflow} and  the same weighted mask $W$ is applied as in the photometric consistency loss.

\textbf{Semantic Consistency Loss.} Photometric loss and 2D correspondence loss may break when the illumination between two images changes drastically. However, despite the illumination changes, $I_2$  and $I_{1\to2}$ should share the same semantic feature representation, as the target image and the rendered image share the same identity, expression and pose. To this end, we minimize the cosine distance between our semantic feature embeddings:

\begin{equation}
L_{s} = \lambda_{s} -\lambda_{s} <\frac {F(I_2)}{||F(I_2)||_2}, \frac {F(I_{1 \to 2})} {||F(I_{1 \to 2})||_2}>,
\label{eqn:semantic}
\end{equation}
where $F$ denotes our shared feature encoder. Unlike existing approaches (\eg, \cite{genova2018unsupervised}) which align semantic features in a pre-trained face recognition network, we simply minimize the feature distance from our learned shared encoder. We find that this speeds up our training process and empirically works better.

\textbf{Regularization Loss.}
Finally, we add a regularization loss to identity and expression parameters to avoid over-fitting:
\begin{equation}
L_{r} = \lambda_{r}\sum_{i=1}^{199} |\frac {\boldsymbol{\alpha}_{id}(i)} {\sigma_{id}(i)} | + \frac{\lambda_{r}}{2}\sum_{i=1}^{29} |\frac {\boldsymbol{\alpha}_{exp}(i)} {\sigma_{exp}(i)} |,
\end{equation}
where $\sigma_{id}$ and $\sigma_{exp}$ represent the standard deviation of $\boldsymbol{\alpha}_{id}$ and $\boldsymbol{\alpha}_{exp}$.

\begin{figure}[t]
    \centering
    \subfigure[2D NME on AFLW2000-3D dataset]{
      \resizebox{0.46\textwidth}{!}{
      \begin{tabular}{l c c c c c c}
         \toprule
           \multirow{2}{*}{Method} & \multicolumn{4}{c}{$NME_{2d}^{68}$}   \\
          & 0 to 30 & 30 to 60 & 60 to 90 & Mean \\
         \hline
         SDM\cite{Xiong2015Global} & 3.67 & 4.94 & 9.67 & 6.12 \\
         3DDFA \cite{zhu2016face} & 3.78 & 4.54 & 7.93 & 5.42\\
         3DDFA + SDM \cite{zhu2016face} & 3.43 & 4.24 & 7.17 & 4.94\\
         Yu et al. \cite{Yu2017Learning} & 3.62 & 6.06 & 9.56 & - \\
         3DSTN\cite{bhagavatula2017faster} & 3.15 & 4.33 & 5.98 & 4.49 \\
         DeFA\cite{liu2017dense} & - & - & - & 4.50\\
         Face2Face~\cite{thies2016face2face} & 3.22 & 8.79 & 19.7 & 10.5\\
         3DFAN~\cite{bulat2017far} & 2.77 & 3.48 & 4.61 & 3.62 \\
         PRN~\cite{feng2018joint} & 2.75 & 3.51 & 4.61 & 3.62\\
         ExpNet~\cite{chang2018expnet} & 4.01 & 5.46 & 6.23 & 5.23 \\
         MMFace-PMN~\cite{yi2019mmface} & 5.05 & 6.23 & 7.05 & 6.11 \\
         MMFace-ICP-128~\cite{yi2019mmface} & 2.61 & 3.65 & 4.43 & 3.56 \\
         \hline
         Ours (PGN) & \textbf{2.49} & \textbf{3.30} & 4.24 & \textbf{3.34} \\
         Ours (3DMM) & 2.53 & 3.32 & \textbf{4.21} &3.36\\
         \toprule
      \end{tabular}
      }
    }
    \quad
    \subfigure[3D NME on AFLW2000-3D dataset]{
      \includegraphics[height=0.36\textwidth]{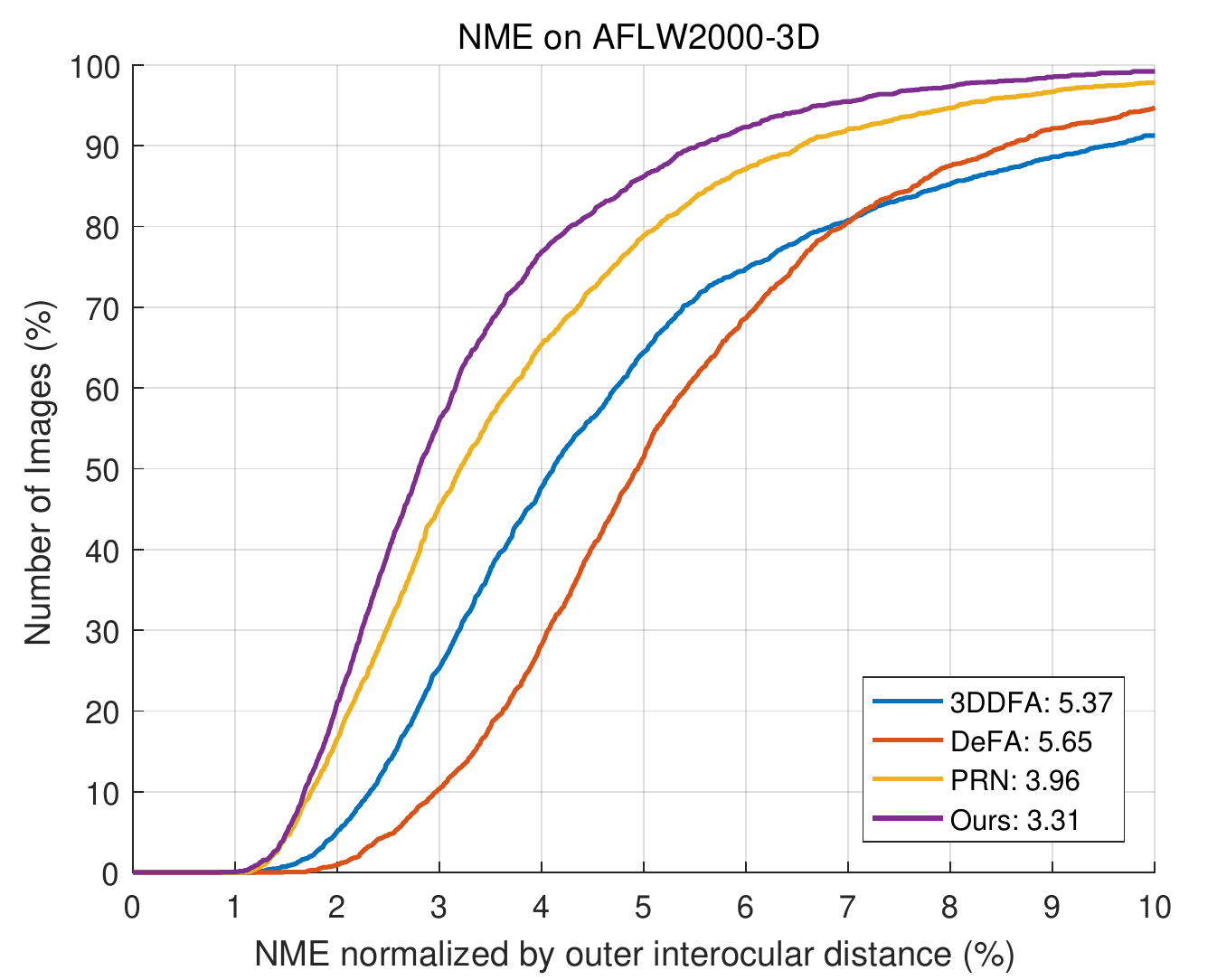}

    }
    \caption{\textbf{Performance comparison on AFLW2000-3D}. (a) \textbf{2D landmarks}. The NME (\%) for 68 2D landmarks with different face orientation along the Y-axis are reported. (b) \textbf{3D face reconstruction}. X-axis denotes the NME normalized by outer interocular distance, the Y-axis denotes the percentage of images. Following~\cite{feng2018joint}, around 45k points are used for evaluation.}
      \label{NME_AFLW}
\end{figure}

\section{Experimental Evaluation}
\textbf{Training Datasets.} To train the shared encoder and PGN, we utilize two types of datasets: synthetic dataset with pseudo 3D annotations and in-the-wild datasets. For synthetic dataset, we choose 300W-LP~\cite{zhu2016face}, which contains 60k synthetic images with fitted 3DMM parameters. These images are synthesized from around 4k face images with face profiling synthetic method~\cite{zollhofer2018state}. To enable more robust 3D face landmark detection, we choose a corpus of in-the-wild datasets, including Menpo~\cite{deng2019menpo}, CelebA~\cite{liu2015faceattributes}, 300-VW~\cite{shen2015first} and Multi-PIE~\cite{gross2010multi} with their 68 2D landmarks automatically extracted by ~\cite{bulat2017far}.

To train identity and expression regression networks with our proposed self-consistency losses, we utilize 300-VW~\cite{shen2015first} and Multi-PIE~\cite{gross2010multi}, where the former contains monocular videos, and the latter contains faces images of the same identity under different lightings, poses, expressions and scenes.

\textbf{Evaluation Datasets and Metrics.} We evaluate our model on AFLW-2000-3D~\cite{zhu2016face}, Florence~\cite{bagdanov2011florence} and FaceWarehouse~\cite{cao2013facewarehouse} datasets. AFLW-2000-3D contains the first 2000 images from AFLW~\cite{koestinger2011annotated}, which is annotated with fitted 3DMM parameters and 68 3D landmarks in the same way as 300W-LP. We evaluate face landmark detection performance and 3D face reconstruction performance on this dataset, which is measured by Normalized Mean Error (NME). Florence dataset contains 53 subjects with ground truth 3D scans, where each subject contains three corresponding videos: ``Indoor-Cooperative", ``PTZ-Indoor" and ``PIZ-Outdoor". We report Point-to-Plane Distance to evaluate 3D shape reconstruction performance. The Florence dataset only contains 3D scans with the neutral expression, which can only be used to evaluate the performance of shape reconstruction. To evaluate the expression part, we further evaluate our method on the FaceWarehouse dataset. Following previous work~\cite{tewari2017mofa,tewari2018self,tran2018nonlinear,tewari2019fml}, we use a subset with 180 meshes (9 identities and 20 expressions each) and report per-vertex error. Florence and FaceWarehouse are also employed to verify the effectiveness of our proposed multi-frame consistency scheme.

\begin{wraptable}{r}{0.5\columnwidth}
   \caption{Comparison of mean point-to-plane error on the Florence dataset.}
  \setlength{\tabcolsep}{3pt}
  \begin{center}
  \resizebox{0.5\textwidth}{!}{
      \begin{tabular}{lcccc}
      \toprule
      \multirow{2}{*}{Method}&
      \multicolumn{2}{c}{Indoor-Cooperative}&
      \multicolumn{2}{c}{PTZ-Indoor}\\
      & Mean & Std & Mean & Std\\
      \midrule
      Tran \etal \cite{tuan2017regressing}
      &1.443&0.292&1.471&0.290\\
      Tran \etal + pool
      &1.397&0.290&1.381&0.322\\
      Tran \etal + \cite{piotraschke2016automated}
      &1.382 & 0.272 & 1.430 & 0.306\\
      MoFA \cite{tewari2017mofa}
      &1.405 & 0.306 &1.306 & 0.261 \\
      MoFA + pool
      &1.370 & 0.321 &1.286 & 0.266 \\
      MoFA + \cite{piotraschke2016automated}
      &1.363  & 0.326 & 1.293 & 0.276\\
      Genova \etal \cite{genova2018unsupervised}
      & 1.405 & 0.339 & 1.271 & 0.293\\
      Genova \etal + pool
      & 1.372 & 0.353 & 1.260 & 0.310\\
      Genova \etal + \cite{piotraschke2016automated}
      & 1.360 & 0.346 & 1.246 & 0.302\\
      MVF~\cite{wu2019mvf} - pretrain & 1.266 & 0.297 & 1.252 & 0.285\\
      MVF~\cite{wu2019mvf} & 1.220 & 0.247 & 1.228 & 0.236\\
      Ours & \textbf{1.122} & \textbf{0.219} & \textbf{1.161} & \textbf{0.224} \\
      \bottomrule
      \end{tabular}
      }
  \end{center}
      \label{table:florence}
\end{wraptable}

\textbf{Training Details.} The face regions are cropped according to either pseudo 3D face landmarks or detected 2D facial landmarks~\cite{bulat2017far}. Then the cropped images are resized to 256$\times$256 as input. The shared encoder and PGN structures are the same as PRN~\cite{feng2018joint}. For PGN, another option is using fully connected layers to regress sparse 3D landmarks, which can reduce a lot of computation with slightly decreased performance. The identity and expression regression networks take the encoder output as input, followed by one convolutional layer, one average pooling layer and three fully-connected layers. 

Our whole training procedure contains 3 steps: (1) We first train the shared encoder and PGN. We randomly sample one batch images from 300W-LP and another batch from in-the-wild datasets, then employ 3D landmark and 2D landmark supervision respectively. We set batch size to 16 and train the network for 600k iterations. After that, both the shared encoder and PGN parameters are fixed. (2) For identity and expression regression networks, we first pre-train them with only one image for each identity as input using $L_{l}$ and $L_{r}$ for 400k iterations. This results in a coarse estimation and speeds up the convergence for training with multiple images. (3) Finally, we sequentially choose 2 and 4 images for each identity as input and train for another 400k iterations by minimizing Eq. (\ref{equ:loss}). The balance weights for loss terms are set to $\lambda_{l}=1$, $\lambda_{p}=0.2$, $\lambda_{f}$ = 0.2, $\lambda_{s}$ = 10, $\lambda_{r}$ = 1. 
Due to the memory consumption brought by rendering and optical flow estimation, we reduce the batch size to 4 for multi-image input. All 3 steps are trained using Adam \cite{kingma2014adam} optimizer with an initial learning rate of $10^{-4}$. Learning rate decays half after 100k iterations.

\textbf{3D Face Alignment Results.} Fig. ~\ref{NME_AFLW}(a) shows the 68 facial landmark detection performance on AFLW2000-3D dataset~\cite{zhu2016face}. 
By training with a large corpus of unlabeled in-the-wild data, our model greatly improves over previous state-of-the-art 3D face alignment methods (\eg, PRN~\cite{feng2018joint}, MMFace~\cite{yi2019mmface}) that heavily rely on 3D annotations. Our method achieves the best performance without any post-processing such as the ICP used in MMFace. Moreover, our PGN is robust. We can fix it and directly use its output as ground truth of 3D landmarks to guide the learning of 3D face reconstruction.

\textbf{Quantitative 3D Face Reconstruction Results.} We evaluate 3D face reconstruction performance with NME on AFLW2000-3D, Point-to-Plane error on Florence
and Per-vertex error on FaceWarehouse. Thanks to the robustness of our PGN, we can directly fix it and obtain accurate pose estimation without further learning. Then, our model can focus more on shape and expression estimation. As shown in Fig.~\ref{NME_AFLW}(b), we achieve the best results on the AFLW2000-3D dataset, reducing $NME_{3d}$ of previous state-of-the-art from 3.96 to 3.31, with 16.4\% relative improvement.

Table~\ref{table:florence} shows the results on the Florence dataset. In contrast to MVF that concatenates encoder features as input to estimate a share identity parameter, we employ average pooling for encoder features, enabling us to perform both single-image and multi-image face reconstruction. In the evaluation setting, it does not make much difference using single-frame or multi-frame as input, because we'll finally average all the video frame output. Notably, our method is more general than the previous state-of-the-art MVF that assumes expressions are the same among multiple images (\ie, multi-view images), while our method can directly train on monocular videos.

\begin{table}[t]
\caption{\textbf{Per-vertex geometric error (measured in mm) on FaceWarehouse dataset.} PGN denotes PGN. Our approach obtains the lowest error, outperforming  the best prior art  \cite{tewari2019fml} by $7.5\%$. 
}
\centering
\resizebox{0.95\textwidth}{!}{
\begin{tabular}{ c c c c c c c c c}
 \toprule
  \multirow{3}{*}{Method} & \multirow{3}{*}{MoFA} & \multirow{3}{*}{Inversefacenet}   & \multirow{3}{*}{ Tewari~\etal} & \multirow{3}{*}{FML}  & Ours & Ours & Ours & Ours\\
   & & & &  & Single-Frame& Single-Frame & Mult-Frame & Multi-Frame\\
   &\cite{tewari2017mofa} & \cite{kim2018inversefacenet} & \cite{tewari2018self} &\cite{tewari2019fml} & without PGN & with PGN & without PGN & with PGN\\

 \midrule

Error & 2.19  & 2.11  & 2.03    & 2.01  & 2.18  & 2.09  & 1.98  & \textbf{1.86}  \\

 \bottomrule
\end{tabular} }
\label{table:FaceWarehouse}
\end{table}

Table~\ref{table:FaceWarehouse} shows the results on FaceWarehouse dataset. For single frame setting,  without modeling albedo, we still achieve comparable performance with MoFA~\cite{tewari2017mofa}, Inversefacenet~\cite{kim2018inversefacenet} and Tewari~\etal~\cite{tewari2018self}. For multi-frame settings, we achieve better results than FML~\cite{tewari2019fml}. For both single-frame and multi-frame settings, we achieve improved performance with PGN. All these show the effectiveness of PGN and self-consistency losses.

\begin{figure}[t]
\centering
\includegraphics[width=0.95\textwidth]{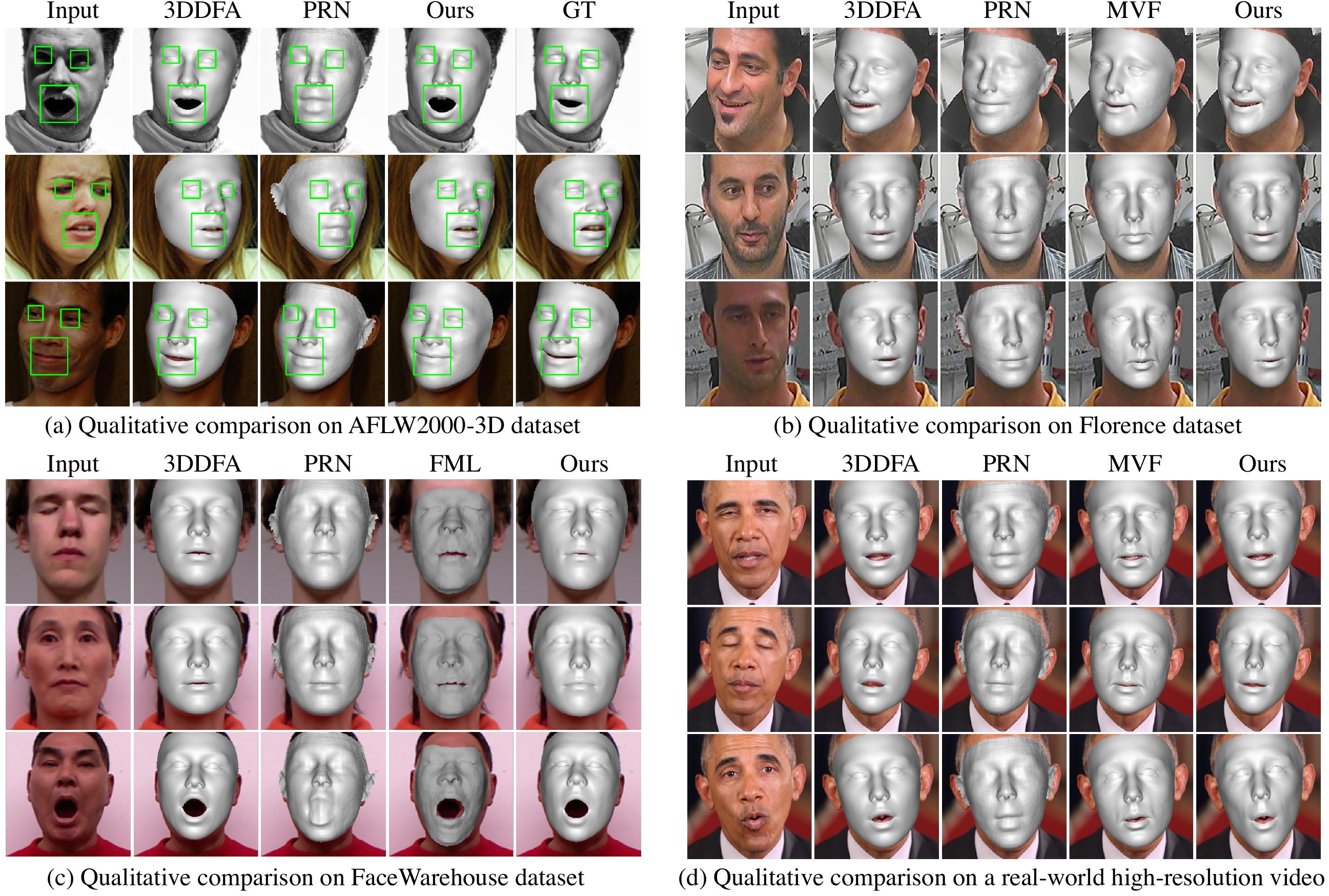}
\caption{\textbf{Qualitative Comparison on various datasets}. Our model generates more accurate shapes and expressions, especially around the mouth and eye region, as we leverage unlimited 2D face data and cross image consistency. The estimated shape of 3DDFA is close to mean face geometry and the results of PRN lack geometric details. (a) On \textbf{AFLW2000-3D}, our results look even more visually convincing than ground truth in many cases. (b) \textbf{Florence}.  (c) \textbf{FaceWarehouse}. Compared with FML, our results are more smooth and visibly pleasing. (d) \textbf{Video results}. Our consistency losses work especially well for high resolution images with few steps of fine-tuning. We generate accurate shape and expression, \eg, challenging expression of complete eye-closing. \textbf{Zoom in for details.} }
\label{Qualitative}
\end{figure}

\textbf{Qualitative 3D Face Reconstruction Results.} Fig.~\ref{Qualitative}(a) shows the qualitative comparisons with 3DDFA~\cite{zhu2016face}, PRNet~\cite{feng2018joint} and the pseudo ground truth. 3DDFA regresses identity, expression and pose parameters together and is only trained with synthetic datasets 300W-LP, leading to performance degradation. The estimated shape and expression of 3DDFA is close to mean face geometry and looks generally similar. PRNet directly regresses all vertices stored in UV position map, which cannot capture the geometric constraints well; thus, it does not look smooth and lacks geometric details, \eg, eye and mouth regions.
In contrast, our estimated shape and expression looks visually convincing. Even when compared with the pseudo ground truth generated with traditional matching methods, our estimation is more accurate in many cases. Fig.~\ref{Qualitative}(b) shows the comparison on the Florence dataset, which further demonstrates the effectiveness of our method. Compared with FML on FaceWarehouse dataset, our results can generate more accurate expressions with visibly pleasing face reconstruction results (Fig.~\ref{Qualitative} (c)).

\begin{figure}[t]
\centering
\includegraphics[width=\textwidth]{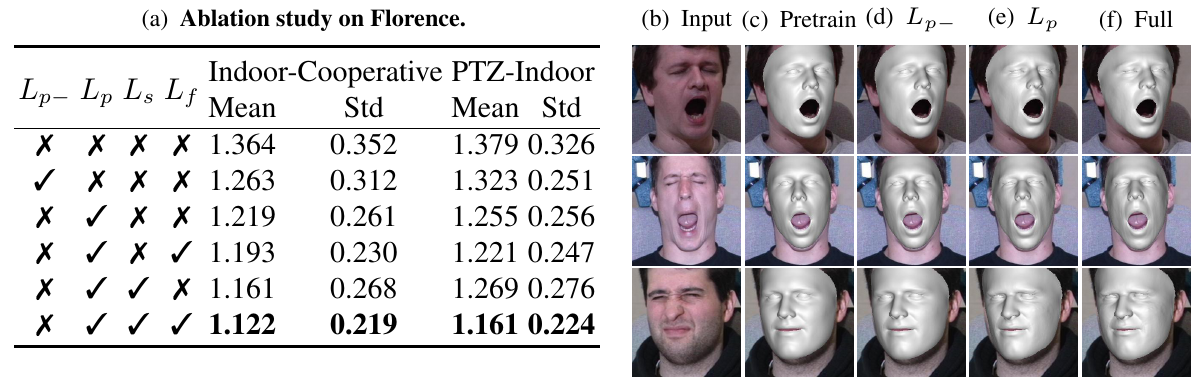}
\caption{\textbf{(a) Ablation Study on Florence dataset.} $L_{p-}$ indicates that census transform is not applied when computing photometric differences. We find that census transform is robust for illumination variations. \textbf{ (b-f) Visual results of our ablations on Multi-PIE dataset:} (b) Input image. (c) Pre-trained model with only landmark loss and regularizer loss. (d) Employ photometric loss. (e) Employ census transform when computing photometric consistency. (f) Full loss. We can find that key components of our model improve the accuracy of shape and expression. \textbf{Zoom in for details.} }
\label{Ablation}
\end{figure}

\textbf{Ablation Study.} The effectiveness of PGN has been shown in Fig~\ref{NME_AFLW} (a) (for face alignment) and Table~\ref{table:FaceWarehouse} (for face reconstruction). To better elaborate the contributions of different components in our self-consistency scheme, we perform detailed ablation study in Fig~\ref{Ablation}.

Our baseline model is single-image face reconstruction trained only with $L_{l}$ and $L_{r}$. However, it doesn't lead to accurate shape estimation, because our PGN with sparse landmarks can only provide a coarse shape estimation. To better estimate the shape, we employ multi-frame images as input. As shown in Fig. ~\ref{Ablation}(a), even without census transform, the photometric consistency ($L_{p-}$) improves the performance. However, photometric loss does not work well when illumination changes among video frames. Therefore, we enhance the photometric loss with census transform to make the model more robust to illumination change. This improves the performance quantitatively (Fig. ~\ref{Ablation}(a)), and qualitatively (Fig.~\ref{Ablation}(b-f)). Applying semantic consistency ($L_{s}$) and flow consistency ($L_{f}$) enforces the rendered image and the target image to look semantically similar and generates better face geometry.

\textbf{Video Results.}
Our proposed multi-image face reconstruction method is based on texture sampling, then it shall obtain better face reconstruction results with higher texture quality (higher video resolution). To verify it, we fine-tune our model on a high-quality video from the Internet, \ie, the fine-tuned model is specialized for the video. No 3D ground truth is used here. As shown in Fig.~\ref{Qualitative}(d), our estimated shape and expression look surprisingly accurate after several thousand iterations. Specifically, our model captures the detailed expression (\eg, totally closed eyes) and face shape very well. This can be an interesting application when we need to obtain accurate 3D face reconstruction for one specific person.

\section{Conclusion}
We have presented a pose guidance network which yields superior performance on 3D face reconstruction from a single image or multiple frames. Our approach effectively makes use of in-the-wild unlabeled images  and provides accurate 3D landmarks as an intermediate supervision to help reconstruct 3D faces. Furthermore, we have demonstrated that swapping textures of multiple images and exploring their photometric and semantic consistency greatly improve the final performance. We hope that our work can inspire future research to develop new techniques that leverage informative intermediate representations (\eg, 3D landmarks in this paper) and learn from unlabeled images or videos.

\section*{Acknowledgement}
This work was partially supported by the RRC of the Hong Kong Special Administrative Region (No. CUHK 14210717 of the General Research Fund) and National Key Research and Development Program of China (No. 2018AAA0100204). We also thank Yao Feng, Feng Liu and Ayush Tewari for kind help.



\bibliographystyle{splncs}
\bibliography{egbib}

\end{document}